# K-Nearest Neighbour and Support Vector Machine Hybrid Classification

**A. M. Hafiz[1]**

[1]Department of Electrical and Electronics Engineering,
Institute of Technology *at* University of Kashmir, Zakura Campus,
Srinagar, J&K 190006, India;
Email: mueedhafiz@uok.edu.in

**ABSTRACT**

*In this paper, a novel K-Nearest Neighbour and Support Vector Machine hybrid classification technique has been proposed that is simple and robust. It is based on the concept of discriminative nearest neighbourhood classification. The technique consists of using K-Nearest Neighbour Classification for test samples satisfying a proximity condition. The patterns which do not pass the proximity condition are separated. This is followed by sifting the training set for a fixed number of patterns for every class which are closest to each separated test pattern respectively, based on the Euclidean distance metric. Subsequently, for every separated test sample, a Support Vector Machine is trained on the sifted training set patterns associated with it, and classification for the test sample is done. The proposed technique has been compared to the state of art in this research area. Three datasets viz. the United States Postal Service (USPS) Handwritten Digit Dataset, MNIST Dataset, and an Arabic numeral dataset – the Modified Arabic Digits Database (MADB), have been used to evaluate the performance of the algorithm. The algorithm generally outperforms the other algorithms with which it has been compared.*

## 1. INTRODUCTION

For improving classification accuracy, hybrid classifiers are used [1-3]. This usually involves obtaining confidence estimates of the classifiers and using this knowledge to make decisions about the combination(s) used. K Nearest Neighbor (KNN) [4] and Support Vector Machine (SVM) [5-8] have been used frequently. KNN-SVM Hybrid Classification is not new [9-16]. There exist many variants of KNN-SVM Hybrid Classifier [9, 17-23]. A popular technique which comes close to our technique is that of Zhang et. al. (2006) [24]. The latter involves KNN classification first, followed by SVM Classification. It generally outperforms KNN and SVM, and is simple. The technique involves performing KNN Classification on those samples for whom first M (usually 50) neighbors are of same class. Next, those samples for which first M neighbors are not same are passed to SVM for classification. For SVM classification of each sample, the SVM is trained on its first P (usually 50) neighbors in neighborhood of overall training set. To make the classification better and faster, pre-computed kernels are used for the SVMs. However, the SVM classifiers trained using this approach, are trained only on first P neighbors (usually 50) which is a small number. Further, the class distribution of these first P neighbors may be random. The performance of discriminative nearest neighbor classification technique can be improved by increasing the number of training samples of its SVMs intelligently. Our technique uses first P neighbors of each class rather than first P neighbors of the overall training set for SVM training. This leads to consistently better results. Further instead of using a large number of neighbors for discrimination, our investigations reveal than a fraction of the same suffices.

Our technique has outperformed the Zhang et. al. (2006) algorithm on all the datasets used. The datasets used include MNIST, USPS and MADB. MNIST and USPS dataset accuracy has also been reported by Zhang et. al. (2006) for their own comparisons. Our technique outperforms SVM on USPS and MADB in terms of accuracy and speed, and its performance is nearly equal to that of SVM on MNIST. It should be noted that using SVM classifier alone for larger number of features on the entire dataset will take much more time and computational resources than the proposed technique. Zhang et. al (2006) have not tested SVM on MNIST, hence we rely on our implementation of their algorithm. On MNIST, SVM performs best, followed by our algorithm, followed by Zhang. et. al (2006), and then by KNN.

The rest of the paper is organized as follows: in section 2, our technique is described in detail; section 3 describes the features used for classification of the data, section 4 shows the performance of the proposed technique on various datasets; the conclusion is given in section 5.

## 2. Proposed Technique

Our technique is based on the concept of discriminative nearest neighborhood classification. It uses KNN and SVM classifiers. The term 'Sifting' refers to the extraction of the closest training patterns for certain test patterns, when the latter fail the proximity condition. The proximity condition is straightforward viz. similarity of first M training samples with the sample under test. As against Zhang. et. al.'s technique, where M is large (usually 50), we obtain good results on a much smaller value of M (between 2 and 6). The advantage of this different approach is that a much smaller number of testing samples are passed to SVM in our algorithm, as compared that of Zhang et. al. KNN is used to classify the 'easier to classify' samples and SVM is used to classify the 'harder to classify' samples. If the algorithm relies too much on KNN, then smaller number of samples are passed to SVM, and vice versa. The main difference between Zhang et. al.'s algorithm and ours is that, the former uses P training samples (50 to 100) for SVM classification whereas our algorithm uses P training samples of every class $C_i$ for SVM classification. The total number of classes 'C' is 10 for digit classification. Here digit classification is used as an example. Hence, Zhang et. al. use P training patterns for each SVM, whereas we use P×C training patterns for each SVM. The advantage is obvious. In our case, the SVMs are better trained to differentiate between various classes as compared to Zhang. et. al.'s algorithm where the first P neighbors may be randomly distributed.

For implementing SVM classification, LIBSVM [25] has been used with Radial Basis Function (RBF) SVMs. We use precomputed kernels [26] for the SVMs, as has been done by Zhang et. al. A Radial Basis Function (RBF) has been used to pre-compute the SVM kernels for both the Zhang et. al. technique and our technique. The use of precomputed kernels leads to faster and more accurate SVM classification by the KNN-SVM hybrid techniques. The RBF kernel function, K used is given by Equation (1).

$$K(\mathbf{x}_{Train}, \mathbf{x}_{Test}) = exp(-\gamma ||\mathbf{x}_{Train} - \mathbf{x}_{Test}||^2), \quad \gamma = 2 \times 10^{-4} \quad (1)$$

where $\mathbf{x}_{Train}$ and $\mathbf{x}_{Test}$ refer to the training and testing set instances respectively.

It should be noted that Zhang et. al. used a simpler kernel function for pre-computation, albeit they state in their work that considered also using RBF Kernels.

Our Algorithm may thus be summarized as follows:
1. For a Training set of R samples and a testing set of S samples, find a pairwise Euclidean distance matrix (S-by-R).
2. For the samples in S, whose first M neighbors are same, use KNN Classification.
3. For the samples in S for whom all M neighbors are not same, set aside P nearest neighbors from each Class in R. Let the number of classes be denoted by C.
4. Train a Multi-class SVM on the P×C samples from R for each sample passed to SVM.
5. Predict the label of each sample (passed to SVM) using the SVM trained in step 4.

### 3. Features Used

Gradient features have been used for the digit images. Two types of gradient features have been used. These are discussed below.

### 3.1 Simple Gradient Features

Gradient Features [27] are robust and popular features. We extracted Gradient Features as follows. First, we used Sobel operators for finding gradients of strokes in horizontal and vertical directions respectively in the binary image.

The horizontal Sobel operator used was:

$$Grd_x = \begin{bmatrix} 1 & 2 & 1 \\ 0 & 0 & 0 \\ -1 & -2 & -1 \end{bmatrix}$$

The vertical Sobel operator used was:

$$Grd_y = \begin{bmatrix} 1 & 0 & -1 \\ 2 & 0 & -2 \\ 1 & 0 & -1 \end{bmatrix}$$

Then we found magnitude and phase of the gradient at each pixel as:

$$M_i = \sqrt{x_i^2 + y_i^2} \quad (2)$$

$$P_i = tan^{-1}\left(\frac{y_i}{x_i}\right) \quad (3)$$

where $M_i$ and $P_i$ are magnitude and phase of gradient at $i^{th}$ pixel in binary image respectively ($i = 1,2,3,...S^2$), where S is size of image array. Also, $x_i$ and $y_i$ are horizontal and vertical components of the gradient at $i^{th}$ pixel after taking respective Sobel operators. Next, we assigned 4 axis or 8 directions in plane of image, where every 2 neighboring axis/directions are separated by angle of 45˚. Thus, the directions were straight lines at following angles (taken anticlockwise): 0, 45, 90, 135, 180, 225, 270, 315 degrees. We further found components of the vector $M_i \angle P_i$ along its two neighboring direction vectors. For all possible variations, we thus had 8 features per pixel in binary image (for 8 directions). Thus, the gradient components of the binary image were reduced to 8 gradient planes. Then we divided the image into $N^2$ (e.g.16) equally sized blocks i.e. N horizontal blocks and N vertical blocks. Then we added the gradient components along 8 respective planes for each block of the image. This gave an 8×$N^2$ element feature per image. N depends on the size of the digit image (e.g. N=16).

### 3.2 HOG Features

Histogram of Gradient (HOG) Features [28-30] have also been used extensively. Though these features are efficient, they are high-dimensional. For computation of the features, we used the inbuilt HOG Feature extractor of MATLAB [31] using 4-by-4 cell size. The HOG Features used had 1296 elements per feature.

### 4. Performance on datasets used

For experimentation purposes, MATLAB was used on an Intel CORE i3, 3GB RAM, Windows 7 system. KNN was used with different distance metrics depending on their respective performances. LIBSVM for MATLAB was used. The performance of the datasets used is shown below.

### 4.1 USPS

The USPS+ Data Set [32] was used for testing purposes which is available online. USPS+ dataset is same as USPS, except that a small number of samples were added to the latter to form the former.

USPS+ has 1100 instances for each digit (0, 1, 2, … 9). Each digit image is a 16×16-pixel grayscale image, given as a 256-element array. The array elements take values from 0 to 255, for 256 gray-levels in the grayscale image. We have used the first 900 instances of each digit as training set and the remaining 200 instances as testing set. Number of classes C is 10. USPS test set is rather difficult as evident from the human error rate of 2.5% [33]. We extract simple Gradient Features from the dataset as described in Section 3.1. The performance of various techniques is shown in Table 1.

Table 1: Performance of Various Techniques on USPS+ Dataset

| Technique | KNN Distance Metric Used | Accuracy (%) | Time Taken (sec) | LIBSVM Cache Size (MB) |
|---|---|---|---|---|
| Ours (K=1) (M=3, P=26) (C=0.297, gamma=0.002) | Euclidean | **99.05** | 30.192 | 200 |
| Zhang et. al. (K=1) (M=3, P=50) (C=0.297, gamma=0.002) | Euclidean | 98.80 | 30.806 | 200 |
| SVM (C=0.297, gamma=0.0025) | - | 98.65 | 77.141 | 2000 |
|  |  |  | 820.541 | 200 |
| KNN (K=6) | Euclidean | 98.30 | 6.132 | - |

As is observable from Table 1, our algorithm performs better, in terms of accuracy, in comparison to the other evaluated techniques including Zhang et. al. Further, it should be noted that SVM Classifier when used alone requires a significant amount of memory (See Column 5 of Table 1), without which it is quite slow when used on the entire dataset. Zhang et. al. also performs better than SVM and KNN classifiers. It should be noted that the time taken by our algorithm and Zhang et. al.'s algorithm is almost same. On an SVM Cache size of 200MB, the top two performing algorithms take almost the same time. SVM, even when allotted an SVM Cache size of 2GB, takes approximately 77 seconds to complete.

### 4.2 MADB

The Modified Arabic Digits Database (MADB) is available online [34, 35]. It has been constructed on lines of the MNIST database. It consists of 60000 instances for training and 10000 instances for testing. We use first 600 instances per digit for training and first 100 instances per digit for testing. Hence the training set size used is 6000 and testing set size used in 1000. We use simple Gradient Features as described in Section 3.1. The performance of various techniques is shown in Table 2.

Table 2: Performance of Various Techniques on MADB

| Technique | KNN Distance Metric Used | Accuracy (%) |
|---|---|---|
| Ours (K=4) (M=3, P=45) (C=0.5, gamma=0.002) | Cosine | **98.70** |
| Zhang et. al. (K=4) (M=5, P=40) (C=0.5, gamma=0.002) | Cosine | 98.50 |
| SVM (C=0.5, gamma=0.002) | - | 98.40 |
| KNN (K=1) | Cosine | 98.20 |

As observed from Table 2, our algorithm outperforms other approaches on the MADB dataset. Zhang et. al.'s algorithm comes next to our algorithm in terms of accuracy, followed by SVM and then KNN.

### 4.3 MNIST

The MNIST dataset of handwritten digits [29] consists of 60,000 training instances and 10,000 testing instances and is available online. Each set consists of an equal number of samples drawn from 2 distinct populations: employees of the Census Bureau, and students of high school. Each digit is a 28x28 grayscale image. Some state-of-the-art algorithms achieve less than 1% error rate, among which Convolutional Network based approach achieves an error rate of 0.23% [36]. We have used first 600 instances for each digit for training and first 100 instances for each digit for testing. Hence the training set size is 6000 and the testing set size is 1000. We use HOG features as described in Section 3.2. The performance of various techniques is shown in Table 3.

Table 3: Performance of Various Techniques on MNIST Dataset

| Technique | KNN Distance Metric Used | Accuracy (%) |
|---|---|---|
| Ours (K=1) | Euclidean | 97.20 |

| (M=3, P=18) (C=4.0, gamma=0.0025) | | |
|---|---|---|
| Zhang et. al. (K=1) (M=2, P=99) (C=0.00025, gamma=2.0) | Euclidean | 96.80 |
| SVM (C=2.0, gamma=0.0004) | - | 97.60 |
| KNN (K=3) | Euclidean | 96.40 |

On MNIST dataset, our algorithm performs second best after SVM, by a narrow margin of 0.40%. As mentioned earlier, Zhang et. al. have not tested SVM classifier on the MNIST dataset. Our implementation of their algorithm comes third, followed by KNN.

### 5. Conclusion

A novel KNN-SVM Hybrid Classification technique has been introduced. It is based on the concept of discriminative nearest neighbor classification. The technique involves using KNN to classify samples which pass a proximity condition viz. similarity of nearest M neighbors. Samples which do not pass the proximity condition are each passed to a respective SVM trained on a combination of their first P nearest neighbors for every class in the training set. The neighborhood nearness is evaluated on basis of a Euclidean distance metric. As per experimentation, a small value of M was found to give optimum results. The optimum value of P was also found to be small. The performance of the proposed technique as against those of various other relevant techniques viz. Zhang et. al (2006), SVM and KNN, have been tested on USPS, MADB and MNIST databases respectively. The algorithm generally outperforms other relevant techniques. The comparison of the proposed technique to that of Zhang et. al. (2006) is specific and the comparisons to KNN and SVM are generic. Zhang et. al. (2006) reduces the comparisons made by the underlying technique common approach. Our technique expands the number of comparisons made and hence brings more information into the decision making ability of the classification process. Judging by speed and memory requirements, our algorithm's performance is also comparable to that of the other evaluated algorithms.